\documentclass[sn-apa,iicol]{sn-jnl} 

\usepackage{color}
\usepackage{stmaryrd}
\usepackage{arydshln}
\usepackage{caption}
\usepackage{subcaption}

\usepackage{enumitem}
\setlist{nolistsep}

\definecolor{ioite}{RGB}{98, 54, 255}
\definecolor{ruby}{RGB}{224, 32, 32}
\definecolor{greycol}{RGB}{230, 230, 230}
\definecolor{ametyst}{RGB}{182, 32, 224}
\definecolor{pink}{RGB}{213, 0, 115}
\definecolor{amber}{RGB}{250, 100, 0}
\definecolor{amazonite}{RGB}{68, 215, 182}
\definecolor{graphite}{RGB}{109, 114,120}
\definecolor{lightgrey}{RGB}{252, 252, 252}
\definecolor{lgreen}{HTML}{5fcf5f} 
\definecolor{lred}{HTML}{db6063}

\lstdefinestyle{blockPython}{
    language=Python,
    frame=tblr,
    commentstyle=\color{amazonite},
    basicstyle=\ttfamily\footnotesize,
    morekeywords={with, as},
    deletekeywords={reduce},
    keywordstyle=\color{pink},
    stringstyle=\color{amber},
    breakatwhitespace=false,         
    breaklines=true,                 
    captionpos=b,                    
    keepspaces=true,                 
    numbersep=5pt,                  
    showspaces=false,                
    showstringspaces=false,
    showtabs=false,                  
    tabsize=2,
    literate=
            *
            {=}{{\textcolor{pink}{=}}}{1}
            {+}{{\textcolor{pink}{+}}}{1}
            {-}{{\textcolor{pink}{-}}}{1}
            {\%}{{\textcolor{pink}{\%}}}{1}
            ,
    emph={init_state, forward, forward_step, forward_steps, __call__, clean_state, load_state_dict, state_dict, randn, allclose, equal, Conv3d,Conv1d, Conv2d, Conv3d, AvgPool1d, AvgPool2d, AvgPool3d, MaxPool1d, MaxPool2d, MaxPool3d, AdaptiveAvgPool1d, AdaptiveAvgPool2d, AdaptiveAvgPool3d, AdaptiveMaxPool1d, AdaptiveMaxPool2d, AdaptiveMaxPool3d, Linear, MultiheadAttention, TransformerEncoderLayer, TransformerEncoder, Sequential, Broadcast, Parallel, ParallelDispatch, Reduce, Residual, Lambda, Delay, Reshape, Add, Multiply, Unity, Constant, Zero, One, LSTM, GRU, Threshold, ReLU, RReLU, Hardtanh, ReLU6, Sigmoid, Hardsigmoid, Tanh, SiLU, Hardswish, ELU, CELU, SELU, GLU, GELU, Hardshrink, LeakyReLU, LogSigmoid, Softplus, Softshrink, PReLU, Softsign, Tanhshrink, Softmin, Softmax, Softmax2d, LogSoftmax, BatchNorm1d, BatchNorm2d, BatchNorm3d, LayerNorm, Dropout, Dropout2d, Dropout3d, AlphaDropout, FeatureAlphaDropout, RNN, RNNCell, LSTMCell, GRUCell, Module, CircularPositionalEncoding, BroadcastReduce},
    emphstyle={\bfseries\color{ioite}}
}

\lstdefinestyle{inlinePython}{
    basicstyle=\ttfamily\normalshape\small,
}

\newcommand{\code}[1]{\mbox{\lstinline[style=inlinePython]{#1}}}
\newcommand{\co}[0]{\code{co} }
\newcommand{\nn}[0]{\code{nn} }

\usepackage{amsthm}
\newtheorem{principle}{Principle}
\newtheorem{assumption}{Assumption}
\newtheorem{definition}{Definition}

\usepackage[capitalize]{cleveref}
\crefname{listing}{Example}{Examples}
\Crefname{listing}{Example}{Examples}
\crefname{section}{Sec.}{Secs.}
\Crefname{section}{Section}{Sections}
\Crefname{table}{Table}{Tables}
\crefname{table}{Tab.}{Tabs.}
\Crefname{principle}{Principle}{Principles}
\crefname{principle}{Pr.}{Prs.}
\Crefname{assumption}{Assumption}{Assumptions}
\crefname{assumption}{As.}{Ass.}
\Crefname{definition}{Definition}{Definitions}
\crefname{definition}{Def.}{Defs.}

\jyear{2022}%

\begin{document}

\title[Continual Inference]{Continual Inference: A Library for Efficient Online Inference with Deep Neural Networks in PyTorch}


\author*[1]{\fnm{Lukas} \sur{Hedegaard}}\email{lhm@ece.au.dk} 

\author[1]{\fnm{Alexandros} \sur{Iosifidis}}\email{ai@ece.au.dk} 

\affil[1]{
\orgdiv{Department of Electrical and Computer Engineering}, 
\orgname{Aarhus University}, 
\orgaddress{
\country{Denmark}}}

\abstract{
We present Continual Inference, a Python library for implementing Continual Inference Networks (CINs) in PyTorch, a class of Neural Networks designed specifically for efficient inference in both online and batch processing scenarios.
We offer a comprehensive introduction and guide to CINs and their implementation in practice, and provide best-practices and code examples for composing complex modules for modern Deep Learning.
Continual Inference is readily downloadable via the Python Package Index and at \url{www.github.com/lukashedegaard/continual-inference}.
    \\
    \paragraph*{Keywords} Online Inference · Continual Inference Network · Deep Neural Network · Python \hfill
}

\maketitle

\section{Introduction}
Designing and implementing Deep Neural Networks (DNNs), which offer good performance in online inference scenarios, is an important but overlooked discipline in Deep Learning and Computer Vision. 
Commonly, research in areas such as Human Activity Recognition focuses heavily on improving accuracy on select benchmark datasets with limited focus on computational complexity and still less on efficient online inference capabilities.
Yet, important real-life applications such as human monitoring~\citep{pigou2018, tavakolian2019}, driver assistance ~\citep{enkelmann2001}, and autonomous vehicles depend on performing predictions on a continual input stream with low latency and low energy consumption. 
This paper is a comprehensive introduction to Continual Inference Networks, the guiding principles of their design, and the Continual Inference Python library for implementing them in PyTorch. 

The remainder of the paper starts by giving an introduction to Continual Inference Networks in \cref{sec:cin}, \cref{sec:design} provides a description of design principles, core modules and composition modules in the Continual Inference library, \cref{sec:experiments} summarizes and compares achieved reductions in step-wise computational complexity and memory-usage for projects using the library, and \cref{sec:conclusion} offers a conclusion.
\section{Continual Inference Networks}\label{sec:cin}
Originally introduced in \cite{hedegaard2021continual} and subsequently elaborated in \cite{hedegaard2022cotrans, hedegaard2021costgcn}, Continual Inference Networks denote a variety of Neural Network, which can operate without redundancy during online inference on a continual input stream, as well as offline during batch inference. 
Specifically, CINs comply with the following definition~\citep{hedegaard2022cotrans}:
\begin{definition}[\textbf{Continual Inference Network}]
A Continual Inference Network is a Deep Neural Network, which
\begin{itemize}
    \item is capable of continual step inference without computational redundancy,
    \item is capable of batch inference corresponding to a non-continual Neural Network,
    \item produces identical outputs for batch inference and step inference given identical receptive fields,
    \item uses one set of trainable parameters for both batch and step inference.
\end{itemize}
\end{definition}

Many prior networks can be viewed as CINs, including networks, which perform their task within a single time-step (e.g. object detection and image recognition models), or which inherently processes temporal data step-by-step (e.g. Recurrent Neural Networks such as LSTMs~\citep{hochreiter1997lstm} and GRUs~\citep{cho2014gru}).
Some network types, however, have been inherently limited to batch inference. These include Convolutional Neural Networks (CNNs) with temporal convolutional components (e.g. 3D CNNs), as well as Transformers with tokens spanning the temporal dimension.
While they can in principle be used for online inference, it is an inefficient process, where input steps are assembled to full batches and fed to the network in a sliding window fashion, with many redundant intermediary computations as a result.

While some specialty architectures have been devised to let 3D convolutional network variants make predictions step by step~\citep{singh2019recurrent, kopuklu2020dissected}, and accordingly also qualify as CINs, these were not weight-compatible with regular 3D CNNs. 

Recently, the \textit{Continual} 3D Convolutions introduced in \cite{hedegaard2021continual} changed this. Through a reformulation of the 3D convolution to compute inputs step-by-step rather than spatio-temporally, well-performing 3D CNNs such as X3D~\citep{feichtenhofer2020x3d}, Slow~\citep{feichtenhofer2019slowfast}, and I3D~\citep{carreira2017quo} trained for Trimmed Activity Recognition were re-implemented to execute step-by-step without any re-training.
Likewise, Spatio-temporal Graph Convolutional Networks for Skeleton-based Action Recognition~\citep{yan2018spatial, shi2019two, plizzari2021skeleton}, which originally operated only on batches, were recently transformed to perform step-wise inference as well though a continual formulation of their Spatio-temporal Graph Convolution blocks~\citep{hedegaard2021costgcn}.

Temporal Transformer networks had likewise been restricted to operate on batches until recently, when \citep{hedegaard2022cotrans} introduced two variant of \textit{Continual} Multi-head Attention (\textit{Co}MHA), which are weight-compatible with the original MHA~\citep{vaswani2017attention}, while being able to compute updated outputs for each time step.

With these innovations, many existing DNNs can be transformed to operate efficiently during online inference. 
In general, non-continual networks, which are transformed to continual ones attain reductions in per-step computational complexity in proportion to the temporal receptive field of the network. In some cases, these savings can amount to multiple orders of magnitude~\citep{hedegaard2021costgcn}.
Still, the implementation of Continual Inference Networks with temporal convolutions and Multi-head Attention in frameworks such as PyTorch~\citep{paszke2019pytorch} requires deep knowledge and practical experience with CINs.
With the Continual Inference library described in the next section, we hope to change this.
\section{Library Design}\label{sec:design}

\subsection{Principles}\label{sec:principles}
The fundamental feature of CINs, that networks are flexible and perform well on both online inference and batch inference, is a guiding principle in the design of the Continual Inference library as well.
Accordingly, refactoring existing implementations in pure PyTorch should be straightforward.
Let us adopt the Python import abbreviations \code{import continual as co} and \code{from torch import nn}. The library then follows the following principle to ensure that \co modules can be used as drop-in replacements for \nn modules without behavior change:
\begin{principle}[Compatibility with PyTorch]
\co modules with identical names to \nn modules also have:
\begin{enumerate}
    \item identical \code{forward},
    \item identical model weights,
    \item identical or extended constructors,
    \item identical or extended supporting functions.
\end{enumerate}
\label{prin:identical}
\end{principle}
Before proceeding to the enhanced functionality of \co modules, let us state our assumption to the input format:
\begin{assumption}[Order of input dimensions]
Inputs to \co modules should use the order $(B, C, T, S_1, S_2, ...)$ for multi-step inputs and $(B, C, S_1, S_2, ...)$ for single-step inputs, where $B$ is the batch size, $C$ is the input channel size, $T$ is the temporal size, and $S_n$ are additional optional dimensions.
\label{ass:dimensions}
\end{assumption}

The core difference between Continual Inference Networks and regular networks is their ability to efficiently compute results for each time-step. 
Besides the regular \code{forward} function found in \nn modules, \co modules add multiple call modes that allow for continual inference with a simple interface: 
\begin{principle}[Call modes]
\co modules provide three forward operations:
\begin{enumerate}
    \item \code{forward}: takes a (spatio-) temporal input and operates identically to the \code{forward} of an \nn module,
    \item \code{forward_step}: takes a single time-step as input without a time-dimension and produces an output corresponding to \code{forward}, had it's input been shifted by one time-step, given identical prior inputs. 
    \item \code{forward_steps}: takes multiple time-steps as input and produces outputs identical to applying \code{forward_step} the number of times corresponding to the temporal size of the input.
\end{enumerate}
Furthermore, the \code{__call__} method of \co modules can be changed to use any of the three by either setting the \code{call_mode} attribute of the module or applying the \code{co.call_mode()} context with a string spelling out the wanted forward type.
\label{prin:forward-modes}
\end{principle}

To get a better understanding of \Cref{prin:forward-modes} in practice, we refer the reader to the examples in Example \ref{code:forward_examples} and \ref{code:call_mode}. \cref{code:forward_examples} shows how the different forward functions introduced in part one of \Cref{prin:forward-modes} can be used.
Part two of \Cref{prin:forward-modes} is exemplified in \cref{code:call_mode}.

Continual modules, which use information from multiple time-steps in their operation, are inherently stateful. 
Whenever \code{forward_step} or \code{forward_steps} is invoked, intermediary results needed for future step results are optimistically computed and stored. 
\Cref{prin:state} summarizes the rules for state-manipulation and updates.

\begin{principle}[State]
Module state is updated according to the following rules:
\begin{itemize}
    \item \code{forward_step} and \code{forward_steps} use and update state by default.
    \item Step results may be computed without updating internal state by passing \code{update_state=False} to either \code{forward_step} or \code{forward_steps}.
    \item \code{forward} neither uses nor updates state.
    \item Module state can be wiped by invoking the \code{clean_state()} method.
    \item A module produces non-empty outputs after its has conducted a number of stateful forwards steps corresponding to its \code{delay}.
\end{itemize}
\label{prin:state}
\end{principle}

Regular \nn modules predominantly operate on input batches in an offline setting and do not have a built-in concept of delay.
\co modules on the other hand are designed to operate on time-series. 
Since \co modules often integrate information over multiple time-steps and online operation is causal by nature, they may only produce the output corresponding to a given input after observing additional steps. For instance, a \code{co.Conv1d} module with \code{kernel_size = 3} can only produce an output from the third input step as illustrated in \cref{fig:-no-padding-delay}.
The delay of a module is calculated according to \Cref{prin:delay}:

\begin{figure}
    \centering
    \includegraphics[width=0.45\linewidth]{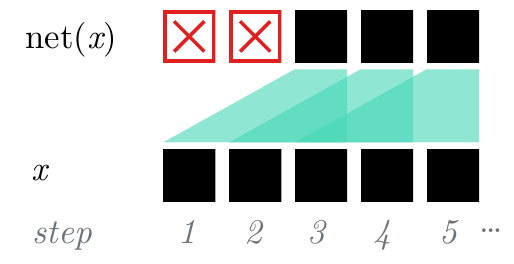}
	\caption{
	    Sketch of delay and receptive field. Here, the step-wise operation of a \co module \code{net} with \code{receptive_field = 3} is illustrated. 
	    $\blacksquare$ are non-zero step-features and \textcolor{ruby}{$\boxtimes$} are empty outputs.
    }
    \label{fig:-no-padding-delay}
\end{figure}

\begin{principle}[Delay]
\co modules produce step outputs that are delayed by
\begin{equation}
    d = f - p - 1
\end{equation}
steps relative to the earliest input step used in the computation, where $f$ is the receptive field and $p$ is the temporal padding.
\label{prin:delay}
\end{principle}

\begin{center}
\vspace{10 pt}
\noindent\begin{minipage}{\columnwidth}
\begin{lstlisting}[
    label={code:forward_examples},
    style=blockPython,
    caption={Definition and usage of \texttt{co.Conv3d} and its forward modes. }
]
import torch
import continual as co
                                                           
con = co.Conv3d(in_channels=4, 
                out_channels=8, 
                kernel_size=3)
assert con.delay == 2
assert con.receptive_field == 3  

reg = torch.nn.Conv3d(in_channels=4, 
                      out_channels=8, 
                      kernel_size=3)
# Reuse weights
con.load_state_dict(reg.state_dict())  

x = torch.randn((2, 3, 5, 6, 7))  # B,C,T,H,W

y = con.forward(x)
z = reg.forward(x)
assert torch.equal(y, z)

# Multiple steps
firsts = con.forward_steps(x[:, :, :4])  
assert torch.allclose(firsts, 
                      y[:, :, : con.delay])
# Single step
last = con.forward_step(x[:, :, 4])  
assert torch.allclose(last, y[:, :, con.delay])
\end{lstlisting}
\end{minipage}
\end{center}

\begin{center}
\vspace{10 pt}
\noindent\begin{minipage}{\columnwidth}
\begin{lstlisting}[
    label={code:call_mode},
    style=blockPython,
    caption={Changing the \texttt{call\_mode} for a continual module \texttt{net}. }
]
net(x)           # Invokes `forward` by default

net.call_mode = "forward_step"
net(x[:, :, 0])  # Invokes `forward_step`

with co.call_mode("forward_steps"):
    net(x)       # Invokes `forward_steps`
    
net(x[:, :, 0])  # Invokes `forward_step` again
\end{lstlisting}
\end{minipage}
\end{center}

\begin{figure}
    \centering
    \includegraphics[width=0.5\linewidth]{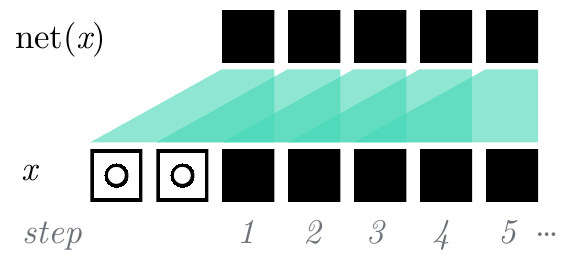}
	\caption{
	    Sketch of how padding reduces delay. Here, the step-wise operation of a \co module \code{net} with \code{receptive_field = 3}, \code{padding = 2} is illustrated. 
	    $\boxcircle$ are padded zeros and $\blacksquare$ are non-zero step-features.
    }
    \label{fig:padding-delay}
\end{figure}

While padding is used in regular networks to retain the size of feature-maps in consecutive layers, this interpretation of temporal padding does not make sense in the context of an infinite, continual input, as handled by CINs.
Instead, we may interpret padding as a reduction in delay. 
For instance, a \code{co.Conv1d} module with \code{kernel_size = 3} and \code{padding = 2} has a delay of zero, because the padded zeros already ``saturated'' the state before-hand. This is illustrated in  \cref{fig:padding-delay}.
Considering, that \co modules expect an infinite continual input stream, end-padding padding is omitted by default.
If an end-padding is required for some reason, it can be applied either by passing manually defined zeros as steps or by setting \code{pad_end = True} for an invocation of the \code{forward_steps} function.

Similar to padding, the stride of a \co module impacts the timing of the outputs. 
Specifically, stride results in empty outputs every $(s-1)/s$ outputs, as well as larger delays for downstream network modules through increased receptive fields. This is stated in Principles \ref{prin:rate} and \ref{prin:receptive-field}.

\begin{principle}[Stride and prediction rate]
For neural network of $N$ modules with strides $s_i, i \in \{1..N\}$, the accumulated stride at any given layer is
\begin{align}
    s_\text{acc}^{(i)} &= s^{(i)} \cdot s_\text{acc}^{(i-1)} \quad i \in {1..N} \label{eq:s_acc} \\
    s_\text{acc}^{(0)} &= s^{(0)}.
\end{align}
Equivalently, the resulting network stride is
\begin{equation}
    s_{NN} = \prod_{i=1}^{N} s^{(i)},
\end{equation}
and the network prediction rate is 
\begin{equation}
    r_{NN} = 1 / s_{NN}.
\end{equation}
Accordingly, the outputs of a \co network are empty every $(s_{NN}-1) / s_{NN}$ steps.
\label{prin:rate}
\end{principle}

\begin{principle}[Accumulated delay]
The accumulated receptive field of a downstream module $i$ in a network of $N$ modules is given by:
\begin{align}
    f_\text{acc}^{(i)} &= f^{(i)} + (f_\text{acc}^{(i-1)} - 1)s^{(i)}, \quad i \in {1..N} \\
    f_\text{acc}^{(0)} &= f^{(0)}.
\end{align}
The accumulated delay of layer $i$ in a network is
\begin{equation}
    d^{(i)} = f_\text{acc}^{(i)} - p_\text{acc}^{(i)} - 1,
\end{equation}
where the accumulated padding $p_\text{acc}$ is given by
\begin{align}
    p_\text{acc}^{(i)} &= p^{(i)} \cdot s_\text{acc}^{(i-1)}, \quad i \in {1..N}, \\
    p_\text{acc}^{(0)} &= p^{(0)}. \label{eq:p_acc_0}
\end{align}
\label{prin:receptive-field}
\end{principle}

\begin{figure}
    \centering
    \includegraphics[width=0.8\linewidth]{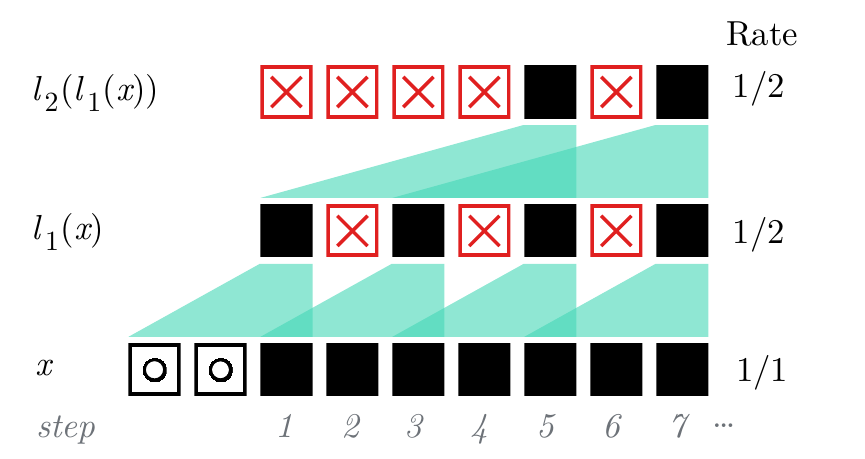}
	\caption{
	    A mixed example of delay and outputs under padding and stride. Here, we illustrate the step-wise operation of two \co module layers, $l_1$ with with \code{receptive_field = 3}, \code{padding = 2}, and \code{stride = 2} and $l_2$ with \code{receptive_field = 3}, no padding and \code{stride = 1}. 
	    $\boxcircle$ denotes a padded zero, $\blacksquare$ is a non-zero step-feature, and \textcolor{ruby}{$\boxtimes$} is an empty output.
    }
    \label{fig:padding-stride-delay}
\end{figure}

\cref{fig:padding-stride-delay} illustrates a mixed example, where the first layer of a two-layer network has \code{padding = 2} and \code{stride = 2}. 
Noting the layer attributes in consecutive order, and using Equations \ref{eq:s_acc} to \ref{eq:p_acc_0}, we have the following network attributes for the example:
\begin{align*}
    s &= \{2,\quad 1\} \\
    p &= \{2, \quad 0\} \\
    s_{acc} &= \{2, \quad 2\cdot1=2\} \\
    p_{acc} &= \{2, \quad 2+2\cdot0=2\} \\
    f_{acc} &= \{3, \quad 3+(3-1)\cdot2 = 7\} \\
    d_{acc} &= \{3-2-1 = 0, \quad 7-2-1 = 4\} \\
    s_{NN} &= s_{acc}^{(1)} = 2 \\
    r_{NN} &= 1 / s_{NN} = 1/2 \\
    d_{NN} &= d_{acc}^{(1)} = 4. \\
\end{align*}
Before continuing onto the specific modules, we have to discuss a final principle of CINs, namely that of parallel modules.
\begin{principle}[Parallel modules]
Modules can be arranged in parallel to execute on each their separate stream of data under the following rules:
\begin{itemize}
    \item Parallel modules follow the same global clock.
    \item The delay of a collection of parallel modules is the maximum delay of any module in the collection.
    \item If the merger of parallel step values includes an empty value, then the resulting step output of the merger is also empty.
\end{itemize}
\label{prin:parallel}
\end{principle}
The intricacies of this principle are best covered with a discussion on residual connections.

\subsubsection{Residual connections}\label{sec:residual}
\begin{figure}[t]
     \centering
     \begin{subfigure}[b]{0.48\linewidth}
         \centering
         \includegraphics[width=0.98\textwidth]{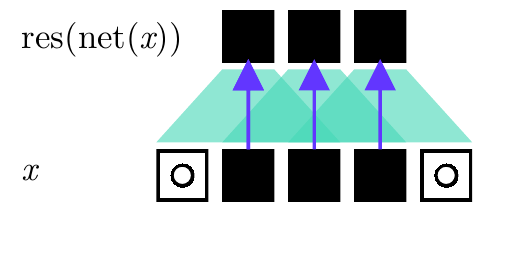}
         \caption{\code{forward}}
         \label{fig:residual-forward}
     \end{subfigure}
     \hfill
     \begin{subfigure}[b]{0.48\linewidth}
         \centering
         \includegraphics[width=0.98\textwidth]{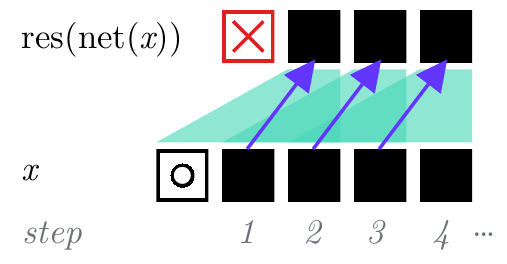}
         \caption{\code{forward_step}}
         \label{fig:residual-step}
     \end{subfigure}
        \caption{Residual connections \textcolor{ioite}{↑} over a module with receptive field of size \textcolor{amazonite}{$\blacktriangle$} and padding one (``equal padding'') $\boxcircle$. \textcolor{ruby}{$\boxtimes$} are empty outputs.}
        \label{fig:residual}
\end{figure}

The residual connection is a simple but crucial tool for avoiding vanishing and exploding gradients in deep neural networks; by adding the input of a module to its output, gradients can flow successfully through models with hundreds of layers.
Without exaggeration, we can state that almost all deep architectures at the time of writing use some form of residual~\citep{he2016resnet, vaswani2017attention, yan2018spatial, feichtenhofer2019slowfast}.
Yet, their implementation in Continual Neural Networks may not follow common intuition in all cases. 
Let us first consider the residual connection during regular \code{forward} operation as found in a non-continual residual shown in \cref{fig:residual-forward}. 
Here, the wrapped module will almost always use padding to ensure equal input and output shapes (known as ``equal padding''). 
For a module with receptive field three, we would thus have a padding of one.
In this case, the \code{forward} computation of the residual simply amounts to adding the input to the output of the convolution.
However, the implementation of \code{forward_step} illustrated in \cref{fig:residual-step} is different.  
Since the first output uses information from the second step, the module has a delay of one. 
Accordingly, the residual connection requires a delay of one as well.

\begin{figure}[thb]
     \centering
     \begin{subfigure}[b]{0.48\linewidth}
         \centering
         \includegraphics[width=0.98\textwidth]{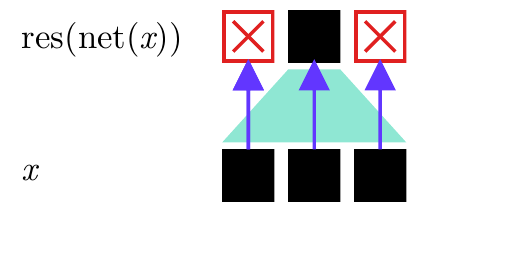}
         \caption{\code{forward}}
         \label{fig:residual-forward-no-pad}
     \end{subfigure}
     \hfill
     \begin{subfigure}[b]{0.48\linewidth}
         \centering
         \includegraphics[width=0.98\textwidth]{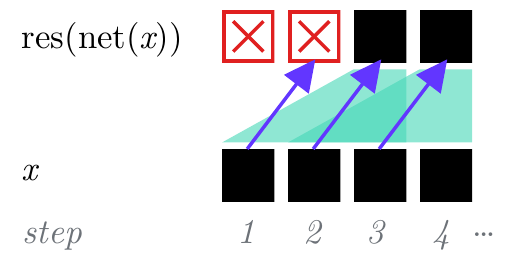}
         \caption{\code{forward_step}}
         \label{fig:residual-step-no-pad}
     \end{subfigure}
        \caption{Centered residual connections \textcolor{ioite}{↑} over a module with receptive field of size \textcolor{amazonite}{$\blacktriangle$} and no padding. \textcolor{ruby}{$\boxtimes$} are empty outputs.}
        \label{fig:residual-no-pad}
\end{figure}

Now consider the same scenario but without padding.
This will be quite foreign to most Deep Learning practitioners, and it is not clear how exactly to align residuals; usually, a separate module would be employed to shrink the residual by an equivalent amount as the wrapped module.
Of the possible alignment choices, a sensible approach is to discard the border values to align the feature maps on \textit{center}.
Contrary to other alignment forms, this has the benefit of weight-compatibility between the no-padding case and the case with equal padding described in the former paragraph; the outputs of step 3 in Figures \ref{fig:residual-step} and \ref{fig:residual-step-no-pad} are equal given the same weights and inputs.
Two issues arise: 
\begin{enumerate}
    \item Delay mismatch: While the residual connection has a delay of one, the wrapped module has a delay of two.
    \item Mix of empty and non-empty results: C.f. the differences in delay, the residual will start producing non-empty outputs before the wrapped module.
\end{enumerate}
\Cref{prin:parallel} helps us navigate this. Despite the internal delay mismatch, the delay of the whole residual module corresponds to the largest delay, in this case two.
Consequently, the whole residual module only produces outputs from the third step, despite the fact that the delayed input already has non-empty outputs from the second step.
Both of these issues can also be avoided if we force residuals to employ the same delay as the wrapped module.
This corresponds to a \textit{lagging} alignment.
However, using such a strategy breaks weight compatibility between the same residual modules with and without padding.

\subsection{Core modules}

Similarly to PyTorch, the Continual Inference library provides a collection of basic building blocks for composing neural networks. 
Following \Cref{prin:identical}, we use the same public interfaces as PyTorch , i.e. class constructor, function names and arguments, and attribute names, to ensure that \co modules can be used as drop-in replacements for \nn modules. 
The basic modules can be categorized as follows:
\begin{itemize}
    \item Convolutions~\citep{hedegaard2021continual}: 
        \code{co.Conv1d}, \code{co.Conv2d}, 
        etc.
    
    \item Pooling: {
        \code{co.AvgPool1d}, 
        \code{co.MaxPool1d}, 
        etc.
    }
    
    \item Linear: \code{co.Linear}.
    
    
    \item Transformer~\citep{hedegaard2022cotrans}: {
        \code{co.MultiheadAttention}, 
        etc.
    }

    \item Shape: 
        \code{co.Delay}, 
        \code{co.Reshape}.
    
    \item Arithmetic:  
        \code{co.Lambda},
        \code{co.Add},
        etc.
\end{itemize}
Here, the \code{co.MultiheadAttention} is a special case, which features two distinct modes of continual operation: 
1)~\code{"single-output"} (default), where only the attention output corresponding to the latest input is produced, and 
2)~\code{"retrospective"}, where updates to prior outputs are also produced retrospectively. The details of this are explained in greater detail in the original paper \citep{hedegaard2022cotrans}.

Linear \co modules follow the \nn modules closely, but ensure compatibility of dimension c.f. \Cref{ass:dimensions}.
\code{co.Delay} adds a specified delay to the input stream. This is handy for aligning the delay of multiple streams as required by residual connections (see \cref{sec:residual}). \code{co.Lambda} allows a user to pass in functions and functors that are applied step-wise to the inputs. 

Besides the above list of tailor-made modules, the Continual Inference library has interoperability with most activation functions (\code{nn.ReLU}, \code{nn.Softmax}, etc.), normalisation layers (\code{nn.BatchNorm1d}, \code{nn.LayerNorm}, etc.), and \code{nn.DropOut} when used within the composition modules as presented in \cref{sec:module-composition}.
        
        
%
This list of compatible module is frequently updated as new innovations find their way into PyTorch.
The current list of compatible modules, is available at \url{www.github.com/lukashedegaard/continual-inference}.

\subsection{Composition modules} \label{sec:module-composition}
In PyTorch, modules are composed by either by using the \code{nn.Sequential} container or by creating a new class which inherits from \code{nn.Module} and manually controls data flow within the \code{forward} function. 
While the latter is commonly used to handle complex modules in a simple and easily debuggable manner, it is not necessarily the simplest approach for implementing complex Continual Inference Networks. 
In addition to defining the basic forward flow, a CIN implementation also needs to handle step-wise computations, which require meticulous alignment of delays if \Cref{prin:forward-modes} is to be kept. 
In practice, this would require three separate forward implementations.

Instead, we expand the container interface of PyTorch to include modules for parallel and conditional processing.
While each module is simple in nature, they can be used to compose complex neural network architectures, which retain all the principles in \cref{sec:principles} without explicitly needing to consider them. 
A brief overview and description of each \co container module is given in \cref{tab:container-modules}.
\begin{table}[b]
	\begin{center}
	\caption{Composition modules.}
	\begin{tabular}{p{0.23\columnwidth}p{0.68\columnwidth}}
		\toprule
        Module  & Description \\
		\midrule
        \code{Sequential}          & Arrange modules sequentially. \\
        \code{Broadcast}           & Broadcast one stream to multiple parallel streams. \\
        \code{Parallel}            & Apply modules in parallel, each on a separate stream. \\
        \code{Reduce}              & Reduce multiple input streams into one. \\
        \code{Residual}            & Add a residual connection for a wrapped module. \\
        \code{Conditional}  & Conditionally invoke a module (or another) at runtime. \\
		\bottomrule
	\end{tabular}
	\label{tab:container-modules}
	\end{center}
\end{table}
To get a practical understanding of these, we will give implementation examples of two common architecture blocks, the residual connection as discussed in \cref{sec:residual} and an Inception module~\citep{szegedy2015going}.

\cref{code:residual_examples} shows three equivalent implementations of a residual 3D convolution block.
\code{res1} is the verbose version, in which \code{co.Broadcast} is used to split a single input into two parallel stream, \code{co.Parallel} specifies that \code{conv} handles the first stream, while a delay is used on the second. 
\code{co.Reduce} merges the streams via an add reduce operation. 
Due to the commonality of broadcast-apply-reduce operations, the library features a \code{co.BroadcastReduce} shorthand to specify such composition more succinctly. 
Even shorter, the \code{co.Residual} module can automatically infer the needed delay from the module it wraps.
Other reduction functions can be specified in \code{co.BroadcastReduce} and \code{co.Residual} using the \texttt{reduce} argument, which is \code{"sum"} by default.
%
\begin{figure}
\vspace{10 pt}
\begin{lstlisting}[
    label={code:residual_examples},
    style=blockPython,
    caption={Equivalent implementations of a residual block. }
]
conv = co.Conv3d(1, 1, kernel_size=3, padding=1)

res1 = co.Sequential(
    co.Broadcast(2),
    co.Parallel(conv, co.Delay(1)),
    co.Reduce("sum"),
)

res2 = co.BroadcastReduce(conv, co.Delay(1))

res3 = co.Residual(conv)
\end{lstlisting}
\vspace{-12pt}
\end{figure}
%
The implementations in \cref{code:residual_examples} correspond to the example in \cref{fig:residual}.
The centered residual module in \cref{fig:residual-no-pad}, which shrinks the residual in \code{forward}, is easily specified as \code{co.Residual(conv, residual_shrink=True)} where \code{conv} has \code{padding = 0}.

We can showcase a more advanced application of parallel streams by considering an Inception module~\citep{szegedy2015going}.
An Inception module broadcasts the input into four stream, and applies convolution of varying kernel sizes in parallel before concatenating the channels to produce one output. 
Without the \co container modules, it would be complicated to keep track of and align delays of the different branches to create valid \code{forward}, \code{forward_step}, and \code{forward_steps} methods.
Using \code{co.Sequential}, which automatically sums up delays, and \code{co.BroadcastReduce}, which automatically adds delays to match the branch with highest inherent delay, the implementation becomes simple as shown in \cref{code:inception_module}.

\begin{figure}
\begin{lstlisting}[
    label={code:inception_module},
    style=blockPython,
    caption={\textit{Continual} Inception module using a mix of \co and \nn modules.}
]
def norm_relu(conv):
    return co.Sequential(
        conv,
        nn.BatchNorm3d(conv.out_channels),
        nn.ReLU(),
    )

inception_module = co.BroadcastReduce(
    co.Conv3d(192, 64, 1),
    co.Sequential(
        norm_relu(co.Conv3d(192, 96, 1)),
        norm_relu(co.Conv3d(96, 128, 3, 
                            padding=1)),
    ),
    co.Sequential(
        norm_relu(co.Conv3d(192, 16, 1)),
        norm_relu(co.Conv3d(16, 32, 5, padding=2))
    ),
    co.Sequential(
        co.MaxPool3d(kernel_size=(1, 3, 3), 
                     padding=(0, 1, 1), 
                     stride=1),
        norm_relu(co.Conv3d(192, 32, 1)),
    ),
    reduce="concat",
)
\end{lstlisting}
\end{figure}

\begin{table*}[!tbp]
\begin{center}
\caption{
    Dataset performance, parameter count, maximum allocated memory (Max mem.), and floating point operations (FLOPs) of continual and non-continual models on video and spatio-temporal graph classification datasets. 
    Subscript$_{xx}$ denotes expanded temporal average pooling, b1 and b2 denote one and two block transformer decoders, and superscript$^*$ indicates architectures where network stride was reduced to one.
    Parentheses indicate the \textcolor{lgreen}{improvement} / \textcolor{lred}{deterioration} of the continual model relative to the corresponding non-continual model. 
    The noted metrics were originally presented in \cite{hedegaard2021continual, hedegaard2022cotrans, hedegaard2021costgcn}.
}
\resizebox{\textwidth}{!}{
\begin{tabular}{lllcll}
    \toprule
    \textbf{Model}
    &\multicolumn{2}{c}{\textbf{Dataset performace} (\%)}
    &\textbf{Params} (M)
    &\textbf{Max mem.} (MB)
    &\textbf{FLOPs} (G)
    \\
    
    \midrule
    & \multicolumn{2}{l}{\textbf{Kinetics-400} 
    (Acc.) } \\
    \cline{2-3}
    X3D-L                             & \multicolumn{2}{l}{69.3}      & \phantom{0}6.2      & 240.7                                           & 19.17  \\
    \textit{Co}X3D-$\text{L}_{64}$    & \multicolumn{2}{l}{71.6 \textcolor{lgreen}{($+2.3$)}}     & \phantom{0}6.2      & 184.4 \phantom{0}\textcolor{lgreen}{($75\%$)}           & \phantom{0}1.25 \phantom{0}\textcolor{lgreen}{($\downarrow15.34\times$)}    \\
    X3D-M                             & \multicolumn{2}{l}{67.2 }     & \phantom{0}3.8      & 126.3                                           & \phantom{0}4.97     \\
    \textit{Co}X3D-$\text{M}_{64}$    & \multicolumn{2}{l}{71.0 \textcolor{lgreen}{($+3.8$)}}     & \phantom{0}3.8      & \phantom{0}69.0 \phantom{0}\textcolor{lgreen}{($55\%$)}  & \phantom{0}0.33 \phantom{0}\textcolor{lgreen}{($\downarrow15.06\times$)}    \\
    X3D-S                             & \multicolumn{2}{l}{64.7 }     & \phantom{0}3.8      & \phantom{0}61.3                                 & \phantom{0}2.06     \\
    \textit{Co}X3D-$\text{S}_{64}$    & \multicolumn{2}{l}{67.3 \textcolor{lgreen}{($+2.6$)}}     & \phantom{0}3.8      & \phantom{0}42.0 \phantom{0}\textcolor{lgreen}{($69\%$)} & \phantom{0}0.17 \phantom{0}\textcolor{lgreen}{($\downarrow12.12\times$)}    \\
    Slow-8×8                          & \multicolumn{2}{l}{67.4 }     & 32.5     & 266.0                                           & 54.87    \\
    \textit{Co}Slow$_{64}$            & \multicolumn{2}{l}{73.1 \textcolor{lgreen}{($+5.7$)}}     & 32.5     & 176.4 \phantom{0}\textcolor{lgreen}{($66\%$)}           & \phantom{0}6.90 \phantom{00}\textcolor{lgreen}{($\downarrow7.95\times$)}    \\
    I3D                               & \multicolumn{2}{l}{64.0 }     & 28.0     & 191.6                                           & 28.61    \\
    \textit{Co}I3D$_{8}$             & \multicolumn{2}{l}{59.6 \textcolor{lred}{($-4.4$)}}       & 28.0     & 235.9 \textcolor{lred}{($123\%$)}          & \phantom{0}5.68 \phantom{00}\textcolor{lgreen}{($\downarrow5.04\times$)}    \\

    \midrule
    
                                    & \multicolumn{1}{l}{\textbf{THUMOS14}
                                    }  
                                    & \multicolumn{1}{l}{\textbf{TVSeries}
                                    }   
                                    \\
                                    & \multicolumn{1}{l}{(mAP)}              & \multicolumn{1}{l}{(mcAP)}  \\
                                    \cline{2-3}
    
    OadTR-b2                        & 64.2                                  & 89.0                           & 15.9     & \phantom{0}67.6      & \phantom{0}1.08 \\ 
	\textit{Co}OadTR-b2             & 64.4 \textcolor{lgreen}{($+0.2$)}     & 88.2 \textcolor{lred}{($-0.8$)}& 15.9     & \phantom{0}71.7 \textcolor{lred}{($106\%$)}      & \phantom{0}0.41 \phantom{00}\textcolor{lgreen}{($\downarrow2.61\times$)} \\ 
	OadTR-b1                        & 64.4                                  & 89.1                           & \phantom{0}9.6      & \phantom{0}43.3      & \phantom{0}0.67 \\ 
	\textit{Co}OadTR-b1             & 64.5 \textcolor{lgreen}{($+0.1$)}     & 88.0 \textcolor{lred}{($-1.1$)}& \phantom{0}9.6      & \phantom{0}45.1 \textcolor{lred}{($104\%$)}      & \phantom{0}0.01 \phantom{0}\textcolor{lgreen}{($\downarrow63.49\times$)} \\ 

    \midrule
    
    & \multicolumn{2}{c}{\textbf{NTU RGB+D 60}
    (Acc.)} \\
    & \textit{X-Sub} & \textit{X-View} \\
    \cline{2-3}
    ST-GCN                               & 86.0  & 93.4  & \phantom{0}3.1      & \phantom{0}45.3    & 16.73 \phantom{ $\downarrow000.0\times$}     \\
    
    \textit{Co}ST-GCN$^*$                & 86.3 \textcolor{lgreen}{($+0.3$)} & 93.8 \textcolor{lgreen}{($+0.4$)}   & \phantom{0}3.1     & \phantom{0}36.1 \phantom{0}\textcolor{lgreen}{($80\%$)}    & \phantom{0}0.16 \textcolor{lgreen}{ ($\downarrow107.7\times$)}     \\
    
    AGCN                                & 86.4  & 94.3  & \phantom{0}3.5   & \phantom{0}48.4  & 18.69 \phantom{ $\downarrow000.0\times$}\\
    
    \textit{Co}AGCN$^*$                 & 84.1 \textcolor{lred}{($-2.3$)}  & 92.6 \textcolor{lred}{($-1.7$)}  & \phantom{0}3.5   & \phantom{0}37.4 \phantom{0}\textcolor{lgreen}{($77\%$)}       & \phantom{0}0.17 \textcolor{lgreen}{($\downarrow108.8\times$)}  \\
    
    S-TR                                & 86.8     & 93.8     & \phantom{0}3.1    & \phantom{0}74.2      & 16.14  \phantom{ $\downarrow000.0\times$}   \\
    
    \textit{Co}S-TR$^*$                 & 86.3 \textcolor{lred}{($-0.3$)}     & 92.4 \textcolor{lred}{($-1.4$)}     & \phantom{0}3.1   & \phantom{0}36.1 \phantom{0}\textcolor{lgreen}{($49\%$)}       & \phantom{0}0.15 \textcolor{lgreen}{($\downarrow107.6\times$)}\\

    \bottomrule
\end{tabular}
}
\end{center}
\label{tab:benchmark}
\end{table*}

\section{Performance comparisons}\label{sec:experiments}

Using the basic \co modules and composition building blocks, continual versions of advanced neural networks have been implemented in multiple recent works with manyfold speedups and significant reductions in memory consumption during online inference~\citep{hedegaard2021continual, hedegaard2022cotrans, hedegaard2021costgcn}. 
Specifically, the 3D-CNNs \textit{Co}X3D, \textit{Co}I3D, and \textit{Co}Slow for video-based Human Activity Recognition were proposed in \cite{hedegaard2021continual}; the Transformer \textit{Co}OadTR for Online Action Detection in \cite{hedegaard2022cotrans}; and Spatio-temporal Graph Convolutional Networks \textit{Co}ST-GCN, \textit{Co}AGCN, and \textit{Co}S-TR for Skeleton-based Action Recognition in \cite{hedegaard2021costgcn}.
While direct conversion from regular to continual versions of the above noted architectures works well in accelerating inference in itself, further improvements can be achieved by exploiting some core characteristics of CINs: in \cite{hedegaard2021continual}, accuracy was improved by increasing model receptive fields through expansions of temporal global average pooling to $64$ steps, and in \cite{hedegaard2021costgcn}, the stride of temporal convolutions was reduced to one to increase prediction rates.
\cref{tab:benchmark} presents a summary of benchmark performance, computational complexity, and maximum allocated memory on GPU for each of these networks alongside with their non-continual counterparts, as presented in prior works~\citep{hedegaard2021continual, hedegaard2022cotrans, hedegaard2021costgcn}.

\section{Conclusion}\label{sec:conclusion}
We presented Continual Inference, an easy-to-use Python library for implementing Continual Inference Networks in PyTorch.
Following interfaces closely, the components provided in the library are backwards-compatible drop-in replacements for PyTorch modules, which add the capability of redundancy-free online inference without the need for intimate knowledge of CINs nor their meticulous low-level implementation. 
Having shown the vast computational advantages of CINs over regular neural networks in multiple settings of video and spatio-temporal graph classification, we hope that this library will contribute to the adoption of CINs and the advancement of use-cases requiring low-latency online inference under recourse constraints in general.

\backmatter


\bmhead{Acknowledgments}
This work has received funding from the European Union’s Horizon 2020 research and innovation programme under grant agreement No 871449 (OpenDR). This publication reflects the authors’ views only. The European Commission is not responsible for any use that may be made of the information it contains.





\bibliography{bibliography}

\end{document}